\documentclass[times,twocolumn,final,authoryear]{elsarticle}
 \biboptions{semicolon,round,sort,authoryear}
\usepackage{prletters}
\usepackage{framed,multirow}

\usepackage{amssymb}
\usepackage{latexsym}

\usepackage{eqnarray}

\usepackage{times}
\usepackage{rotating}
\usepackage[cmex10]{amsmath}
\usepackage{eqnarray}
\usepackage{latexsym} 
\usepackage{wrapfig}
\usepackage{tikz}
\usepackage{subfigure}

\usepackage{url}
\usepackage{xcolor,psfrag}
\definecolor{newcolor}{rgb}{.8,.349,.1}

\journal{Pattern Recognition Letters}

\begin{document}
 \clearpage

\ifpreprint
  \setcounter{page}{1}
\else
  \setcounter{page}{1}
\fi

\begin{frontmatter}

\title{Max-Margin Feature Selection}

\author[1]{Yamuna \snm{Prasad}\corref{cor1}} 
\cortext[cor1]{Corresponding author: 
  Tel.: +91-965-069-0223  
  }
\ead{yprasad@cse.iitd.ac.in}
\author[1]{Dinesh \snm{Khandelwal}}
\author[1]{K. K. \snm{Biswas}}

\address[1]{Department of Computer Science \& Engineering,  
           Indian Institute of Technology Delhi, 
           New Delhi, 110016, India}

\received{1 May 2013}
\finalform{10 May 2013}
\accepted{13 May 2013}
\availableonline{15 May 2013}
\communicated{S. Sarkar}
\begin{abstract}
Many machine learning applications such as in vision, biology
and social networking deal with data in high dimensions. Feature
selection is typically employed to select a subset of features
which improves generalization accuracy as well as reduces the 
computational cost of learning the model. One of the criteria
used for feature selection is to jointly minimize the redundancy
and maximize the relevance of the selected features. In this paper, 
we formulate the task of feature selection as a one class SVM problem 
in a space where features correspond to the data points and instances 
correspond to the dimensions. The goal is to look for a representative 
subset of the features (support vectors) which describes the boundary 
for the region where the set of the features (data points) exists. This 
leads to a joint optimization of relevance and redundancy in a principled 
max-margin framework. Additionally, our formulation enables us to leverage 
existing techniques for optimizing the SVM objective resulting in highly 
computationally efficient solutions for the task of feature selection. 
Specifically, we employ the dual coordinate descent algorithm 
(Hsieh et al., 2008), originally proposed for SVMs, for our formulation. We use a sparse 
representation to deal with data in very high dimensions. Experiments on 
seven publicly available benchmark datasets from a variety of domains show that 
our approach results in orders of magnitude faster solutions even while retaining 
the same level of accuracy compared to the state of the art feature selection 
techniques.
\end{abstract}

\end{frontmatter}
\section{Introduction}
\label{intro}
Many machine learning problems in vision, biology, social networking 
and several other domains need to deal with very high dimensional data. 
Many of these attributes may not be relevant for the final prediction task and act 
as noise during the learning process. 
A number of feature selection methods have already been proposed in the 
literature to deal with this problem. These can be 
broadly categorized into filter based, wrapper 
based and embedded methods.

In filter based methods, features (or subset of the features) 
are ranked based on their statistical importance and are 
oblivious to the classifier being used \citep{guyon&elisseeff03, peng&al05}.
Wrapper based methods select subset of features heuristically and classification 
accuracy is used to estimate the goodness of 
the selected subset \citep{ganesh12}.
These methods typically result in good accuracy while incur 
high computational cost because of the need 
to train the classifier multiple number of times.
In the embedded methods, feature selection criteria is directly incorporated 
in the objective function of the classifier \citep{mingkui&al10, yiteng&al12}.
Many filter and wrapper based methods fail on very
high dimensional datasets due to their high time and memory requirements,
and also because of inapplicability on sparse datasets \citep{guyon&elisseeff03, yiteng&al12}. 

In the literature, various max-margin formulation had been developed for 
many applications \citep{burges98, guo07}.
Recently, we have proposed a hard margin primal formulation 
for feature selection using quadratic program (QP) slover \citep{prasad&al13}. 
This approach jointly minimizes redundancy and maximizes relevance 
in a max-margin framework.
We have formulated the task of feature selection as a one class SVM 
problem \citep{scholkopf&al00} in the dual space where {\emph features} correspond to 
the data points and instances correspond to the dimensions. The goal is to search for a 
representative subset of the features (support vectors) which describes the boundary for
the region in which the set of the features (data points) lies.
This is equivalent to searching for a hyperplane which maximally separates
the data points from the origin \citep{scholkopf&al00}. 
%

In this paper, we have extended the hard-margin formulation to develop a 
general soft-margin framework for feature selection. We have also 
modified the primal and dual formulations. We present the dual objective
as unconstrained optimization problem.
%
%
We employ the Dual Coordinate 
Descent (DCD) algorithm \citep{hsieh&al08} for solving our formulation. The DCD algorithm 
simultaneously uses the information in the primal as well as in the dual to come up with a 
very fast solver for the SVM objective. 
In order to apply DCD approach, our formulation has 
been appropriately modified by including an additional term in the dual objective, which 
can be seen as a regularizer on the feature weights. 
The strength of this regularizer can be tuned to control
the sparsity of the selected features weights. 
We adapt the liblinear implementation \citep{Fan08} for 
our proposed framework so that our approach is 
scalable to data in very high dimensions.
%
We also show 
that  the Quadratic Programming Feature 
Selection (QPFS) \citep{lujan&al10} falls out as a special case of our formulation 
in the dual space when using a hard margin. 

Experiments on seven publicly available datasets from a vision, biology 
and  Natural Language Processing (NLP) domains show that 
our approach results in orders of magnitude faster solutions 
compared to the state of the art techniques while retaining the same 
level of accuracy.

The rest of the paper is organized as follows.
We describe our proposed 
max-margin formulation for feature selection (MMFS) including the dual coordinate 
descent approach in Section \ref{sec:mmfs}. 
We present our experimental evaluation in 
Section \ref{sec:expt}. We conclude our 
work in Section \ref{sec:conc}.
 \vspace{-.12cm}
\section{Proposed Max-Margin Framework\label{sec:mmfs}}
The key objective in feature selection 
is to select a subset of features which are highly relevant (that is high 
predictive accuracy) and non-redundant (that is uncorrelated). Relevance is 
captured either using an explicit metric (such as the correlation between
a feature and the target variable) or implicitly using
the classifier accuracy on the subset of features being
selected. Redundancy is captured using metrics such
as correlation coefficient or mutual information. Most of the 
existing feature selection methods rely on a pairwise
notion of similarity to capture redundancy \citep{lujan&al10, peng&al05, yu&liu03}. 

We try to answer the question "Is there a principled 
approach to jointly capturing the relevance as well redundancy amongst 
the features?". To do this, we flip around the problem and examine 
the space where features themselves become the first class objects.
In particular, we analyze the space where "features" represent
the data points and "instances" represent the dimensions. Which boundary
could describe well the set of features lying in this space? Locating 
the desired boundary is similar to one class 
SVM formulation \citep{scholkopf&al00}.
This equivalently can be formulated as the problem of searching for 
a hyperplane which maximally separates the features (data points) from 
the origin in the appropriate kernel space over the features. In order
to incorporate feature relevance, we construct a set of parallel marginal
hyperplanes, one hyperplane for each feature. The margin of each separating hyperplane
captures the relevance of the corresponding feature. Greater the relevance, 
higher the margin required (a greater margin increases the chances of a 
feature being a support vector). Redundancy among the features is captured 
implicitly in our framework. The support vectors which lie on respective margin 
boundaries constitute the desired subset of features to be selected. This leads
to a principled max-margin framework for feature selection.
%
%
The proposed formulation for MMFS is presented hereafter.
\subsection{Formulation}
Let $X$ represent the data matrix where each row vector ${x_i}^T$ 
($i \in 1 \ldots M)$ denotes an instance and each column 
vector $f_j$ ($j \in 1 \ldots N)$ denotes a feature vector. We will use
$\phi$ to denote a feature map such that the dot product between the data 
points can be computed via a kernel $k(x_i,x_j) = \phi(x_i)^T \phi(x_j)$, 
which can be interpreted as the similarly of $x_i$ and $x_j$. We will
use $Y$ to denote the vector of class labels $y_i$'s $(i \in 1 \ldots M)$.
Based on the above notations, we present the following 
formulation for feature selection
in the primal:
\begin{equation}
\label{eqn:mmfs-primal}
\begin{array}{ll}
\begin{aligned}
{} & \min_{w,b}\quad \frac{1}{2}w^Tw + b + C\sum_{i=1}^N \xi_i\\
\text{subject to} & ~~~~ w^T\phi(f_i) + b \geq r_i - \xi_i, ~ \xi_i \geq 0,~~ \forall i = 1, \ldots, N;
\end{aligned}
\end{array}
\end{equation}
where, $w$ represents a vector normal to the separating hyperplane(s)~\footnote{All the separating
hyperplanes are parallel to each other in our framework.}, $b$ represents the bias
term and $\xi_i$'s represent slack variables. $r_i$ captures the relevance for the $i^{th}$ 
feature. The equation of the separating hyperplane is given by $w^T\phi(f_i)+b=0$ with the 
distance of the hyperplane from the origin being $-b$. 
Note that in this formulation the objective function is similar to the one 
class SVM \citep{scholkopf&al00}.
However, the constraints are very much different as 
our formulation includes the relevance of the features ($r$).
The choice of $\phi$ 
determines the kind of similarity (correlation) to be captured among the features. The set of
support vectors obtained after optimizing this problem i.e. \{$f_i \mid w^T\phi(f_i) + b = r_i$\} and the
margin violators \{$f_i \mid \xi_i > 0$\} constitute the set of features to be selected. In the
dual space, this translates to those features being selected for which $0 < \alpha_i \leq C$ where
$\alpha_i$ is the Lagrange multiplier for $f_i$. 
We will refer to our approach as Max-Margin Feature Selection (MMFS).
Note that when dealing with hard margin (no noise) case and the term 
involving $C$ disappears (since this enforces $\xi_i=0, \forall i$).

Figure 1 illustrates the intuition behind our proposed framework in the 
linear dot product space (with hard margin). In the figure, $w^Tf +b = 0$ 
represents the separating hyperplane. The distance of this hyperplane from the 
origin is given by $-b/||w||$. The first term in the objective of 
Equation \ref{eqn:mmfs-primal} tries to minimize $w^Tw$ i.e. maximize $1/||w||$. 
The second term in the objective tries to minimize $b$ i.e. maximize $-b$. Hence, 
the overall objective tries to push the plane away from the origin. The i$^{th}$ 
dashed plane represents the margin boundary for the i$^{th}$ feature. The distance 
of this marginal hyperplane from the separating hyperplane is given by $r_i/||w||$ 
where $r_i$ is the pre-computed relevance of the i$^{th}$ feature. Therefore, minimizing 
$w^Tw$ in the objective also amounts to maximizing this marginal distance ($r_i/||w||$). 
Hence, the objective has the dual goal of pushing the hyperplane away from the origin 
while maximizing the margin for each feature (weighted by its relevance)as well. The 
features which lie on the respective marginal planes are the support features 
(encircled points). The redundancy is explicitly captured in 
the dual formulation of this problem.
\begin{figure}[!htb]
\label{figPrimal}
\begin{center}
\begin{tikzpicture}[scale=.77]
  \coordinate (A1) at (0,0);
 \coordinate (A2) at (8,0);
  \coordinate (A3) at (0,8);
 
\draw[thick,<->] (A3) -- (A1) --(A2);
\draw[black, thin] (5.15,0)--(0,5.15);
\draw [black, dashed, thin] (6,0) -- (0,6);
\draw [black, dashed, thin] (7.5,0) -- (0,7.5);
\draw [black, dashed, thin] (8.4,1.4) -- (1.8,8);
 \node [rotate=45, very thick] at (2.9,4.0) {$\ldots$};
 
 \node [rotate=45, very thick] at (3.0,5.2) {$\ldots$};
 \node [rotate=45, very thick] at (3.35,5.55) {$\ldots$};
 
 \draw [fill=gray] (1.95,3.95) rectangle (2.05,4.05);
 \draw [black, thin] (2,4) circle [radius=0.15];     
 
 \draw [fill=gray] (3.95,5.75) rectangle (4.05,5.85);
  \draw [black, thin] (4,5.8) circle [radius=0.15];   
  
 \draw [fill=gray] (5.75, 3.95) rectangle (5.85,4.05);
  \draw [black, thin] (5.8,4) circle [radius=0.15];     
  
 \draw [fill=gray] (3.45,3.95) rectangle (3.55,4.05);
  \draw [black, thin] (3.5,4) circle [radius=0.15];    
  
 \draw [fill=gray] (3.95,3.45) rectangle (4.05,3.55);
  \draw [black, thin] (4,3.5) circle [radius=0.15];    
  
 
\draw [fill=gray] (2.3,3.6) rectangle (2.4,3.7);
\draw [black, thin] (2.35,3.65) circle [radius=0.15];   

\draw [fill=gray] (2.4,3.8) rectangle (2.5,3.9);
 
 \draw [fill=gray] (2.5,4.3) rectangle (2.55,4.35);
 \draw [fill=gray] (3.1,4.4) rectangle (3.15,4.45);
 \draw [fill=gray] (4.3,4.4) rectangle (4.35,4.45);
 \draw [fill=gray] (4.3,4.7) rectangle (4.35,4.75);
 \draw [fill=gray] (4.2,3.3) rectangle (4.25,3.35);
 \draw [fill=gray] (3.8, 4.7) rectangle (3.85,4.75);
 \draw [fill=gray] (4.1,2.8) rectangle (4.15,2.85);

   \draw [fill=gray] (3.95,2.15) rectangle (4.0,2.2);
   \draw [fill=gray] (3.3,3.7) rectangle (3.35,3.75);
   \draw [fill=gray] (5.35,3.95) rectangle (5.4,4.0);
   \draw [fill=gray] (4.45,3.95) rectangle (4.5,4.0);
 \draw[black, thin,<->] (0,0)--(2.55,2.55);
\draw[black, thin,<->] (2.6,2.6)--(3.75,3.75);

 \node at (1.5,0.7) {$-b/||w||$};
 \node [rotate=315] at (4.55,2.35) { Margin ($r_i/||w||$) };

 \node [rotate=315] at (3.4,1.4) {$w^T f + b = 0 $};
 \node [rotate=315] at (3.1,1.1) {\small{Separating hyperplane}};

 \node [rotate=315] at (1.3,6.7) {$w^T f + b = r_i $};
 \node [rotate=315] at (1.6,7.0) {\small{i$^{th}$ marginal hyperplane}};

 \draw [thin, ->] (5.5,6.5) -- (4.05, 5.85);
 \draw [thin, ->] (5.5,6.5) -- (5.85, 4.05);
 \draw [thin, ->] (5.5,6.5) -- (3.55, 4.05);
   
 \node [rotate=0] at (5.6, 6.6) {Support Features};
 
 \node [rotate=0] at (4,-.3) {$x_1$};
 \node [rotate=90] at (-0.3,4) {$x_2$};
 \end{tikzpicture}
\end{center}
\caption{Feature representation in sample space. The diagram is conceptual only.}
\end{figure}
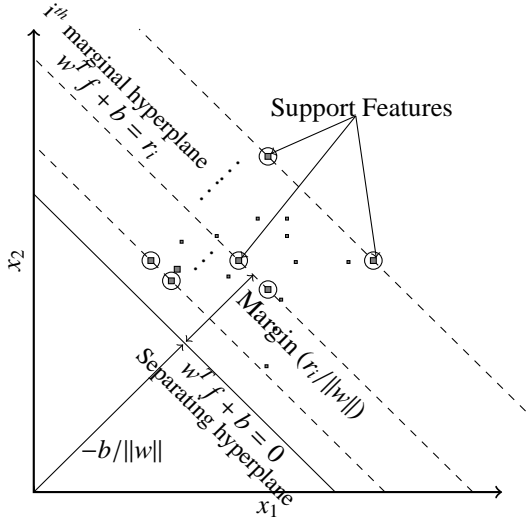
\subsection{Dual Formulation}
In order to solve the MMFS optimization efficiently by Dual Coordinate 
Descent strategy, 
we require both the primal and dual formulations. The dual 
formulation for Equation \ref{eqn:mmfs-primal} can be derived
using the Lagrangian method. The Lagrangian function 
$L(w,b,\xi,\alpha,\beta)$ can be written as:
\begin{equation*}
 \begin{aligned}  
  L(w,b,\xi,\alpha,\beta) &= \min_{w,b} \quad \frac{1}{2}w^Tw + b + C\sum_{i=1}^N \xi_i \\
 {}& + \sum_{i=1}^N \alpha_i (r_i -\xi_i -(w^T\phi(f_i) + b)) - \sum_{i=1}^N \beta_i \xi_i
   \end{aligned}
\end{equation*}
Where, $\alpha_i$'s and $\beta_i$'s are the Lagrange multipliers. Now, 
the Lagrangian dual can be written as:
\begin{equation}
\label{mmfs:max-min}
 \begin{aligned}
  \max_{\alpha,\beta: \alpha_i \geq 0, \beta_i \geq 0}\quad \min_{w,b,\xi} \quad L(w,b,\xi,\alpha,\beta) 
  \end{aligned}
\end{equation}
At the optimality, $\nabla_{w}L$, $\frac{\partial L}{\partial b}$ and $
\frac{\partial L}{\partial \xi_i}$ (for all $i$) will be 0 i.e.
\begin{equation}
\label{mmfs:kkt}
 \begin{aligned}
 {} &\nabla_{w} L  = w - \sum_{i=1}^N \alpha_i \phi(f_i)  = 0 ;
 \frac{\partial L}{\partial b}  = 1 -\sum_{i=1}^N \alpha_i =0 \\
 {} &\frac{\partial L}{\partial \xi_i}  = C - \alpha_i - \beta_i  = 0 
  \end{aligned}
\end{equation}
By substituting the values from Equation \eqref{mmfs:kkt} into Equation \eqref{mmfs:max-min} we get:
\begin{equation}
\label{eqn:mmfs-maxdual}
\begin{aligned}
 f\left(\alpha\right) & = \max_{\alpha} \quad r^T\alpha - \frac{1}{2}\alpha^TQ\alpha\\
 \text{Subject to}~~~ & 0 \leq \alpha_i \leq C,  i = 1, ..., M; ~I^T\alpha = 1. 
\end{aligned}
\end{equation} 
This is similar to the standard SVM dual derivation \citep{scholkopf&al00}.
The only difference is that while there is a single margin 
in standard SVM, the number of features here dictate the number of margins .
We can equivalently rewrite the dual 
formulation of \eqref{eqn:mmfs-maxdual}
as follows:
\begin{equation}
\label{eqn:mmfs-dual}
\begin{aligned}
 f\left(\alpha\right) & = \min_{\alpha} \quad \frac{1}{2}\alpha^TQ\alpha - r^T\alpha \\
 \text{Subject to} ~~~& 0 \leq \alpha_i \leq C,  i = 1, ..., M; ~ I^T\alpha = 1.
\end{aligned}
\end{equation} 
Here, $Q$ is the similarity matrix whose entries are given
by $Q_{ij}=k(f_i,f_j)$ where $k(f_i,f_j)=\phi(f_i)^{T}\phi(f_j)$ 
is the kernel function corresponding to the dot product in the transformed 
feature space. $r$ represents the vector of feature relevance. $\alpha$'s are 
the Lagrange multipliers. Note that the first term in the objective captures the 
redundancy between the features and the second term captures the relevance as 
in the case of QPFS formulation of \citep{lujan&al10}. 
Hence, the connection between the redundancy and the relevance becomes explicit in the
dual formulation. It should be noted that the dual objective bears a 
close similarity to the QPFS objective. 
We give the detailed comparison in Section \ref{sec:related}. We can give relative 
importance to redundancy and relevance by incorporating a scaling parameter 
$\theta \in (0,1)$ in Equation \eqref{eqn:mmfs-dual} as follows:
\begin{equation}
\label{eqn:mmfs-dual2}
\begin{aligned}
 f\left(\alpha\right) & = \min_{\alpha} \quad \frac{1}{2} (1- \theta) \alpha^TQ\alpha - \theta r^T\alpha \\
 \text{Subject to}~~~ & 0 \leq \alpha_i \leq C,  i = 1, ..., M; ~ I^T\alpha = 1.
\end{aligned}
\end{equation}
In the primal formulation (Equation \eqref{eqn:mmfs-primal}), this can be achieved by 
scaling the relevance scores by $\frac{\theta}{1-\theta}$, that is, replacing the 
constraints $ w^T\phi(f_i) + b \geq r_i - \xi_i$ by $ w^T\phi(f_i) + b \geq 
\frac{\theta}{1-\theta} (r_i - \xi_i)$.
\subsection{Choice of Metrics}
The relevance of a feature in our framework is captured using the 
correlation between the feature vector and the class label vector. In our 
experiments, we have normalized the data 
as well as the target vector (class labels) so 
that it has zero mean and unit variance. Hence, the dot product between the 
feature vector and the target vector (normalized) estimates the correlation 
between them i.e. relevance of the $i^{th}$ feature can be computed as 
$r_i = |Y^T \phi (f_i)|$. Some other appropriate metric which captures the 
predictive accuracy of a feature (such as mutual information(MI)) could also 
be used \citep{peng&al05}.

The redundancy is usually captured using correlation or mutual information in 
feature selection tasks \citep{peng&al05}. In our framework, the dot product 
space (kernel) captures the similarity (redundancy) among the 
features and the required similarity metric can be captured by selecting 
the appropriate kernel. The linear kernel ($f_i^Tf_j$) represents the correlation 
among the features when the features are normalized to zero mean and unit variance
{\footnote {It is typical to normalize the data to zero mean and unit variance for
feature selection.}}. 
Since the value of the correlation ranges between $-1$ and $1$, 
a degree two homogeneous polynomial kernel defined over normalized data represents 
the squared correlation (i.e. ${\phi (f_i)}^T {\phi (f_j)} = (f_i^Tf_j)^2$).
The choice of this kernel is quite intuitive for feature selection as it 
gives equal importance to the positive and negative correlations. A Gaussian kernel 
can also be used to approximate the mutual information (MI) \citep{Gretton05} which 
is the key metric for non-linear redundancy measure in feature selection 
problems \citep{peng&al05, lujan&al10}.
%
%
%
Since the MMFS formulation very closely matches the one class SVM 
formulation, any of the existing algorithms for SVM optimization either 
in primal or dual can be used. Next, we describe the use of Dual Coordinate 
Descent (DCD) algorithm \citep{hsieh&al08} to obtain a highly 
computationally efficient solution for our feature 
selection formulation.
\subsection{Dual Coordinate Descent for MMFS}\label{sec:mmfs-dcd}
Following equation \eqref{eqn:mmfs-primal}, the number of variables 
and the number of constraints in the primal formulation 
are $M$+$1$ and $2N$, respectively, while 
from equation \eqref{eqn:mmfs-dual2}, it 
is seen that the corresponding numbers are $N$ and $2N$+$1$, respectively.
Solving the primal (typically by using QP solvers) may be efficient ($O(M^3)$)
in the cases when $M\ll N$ \citep{shalev&al07}.
Solving the dual using QP solvers requires $O(N^2)$ 
space and $O(N^3)$ time. Even solving the dual using 
sequential minimal optimization (SMO) based methods in practice 
has the complexity of $O(N^2)$ \citep{Fan05}.
These high {\emph time} 
and {\emph memory complexities}
limit the scalability of directly solving the primal or dual for data with
a very large number of instances and features.

In many cases when the data already lies in a rich feature space, the
performance of linear SVMs is observed to be similar to that of 
non-linear SVMs. In such scenarios, it may be much more efficient to train 
the linear SVMs directly. The dual coordinate descent methods have been well 
studied for solving linear SVMs using unconstrained form of the primal 
as well as dual formulations \citep{hsieh&al08} who 
have shown that dual coordinate descent algorithm is significantly faster than 
many other existing algorithms for solving the SVM problem. Since our formulation very 
closely resembles the one class SVM formulation (with the exception of having a separate
margin for each feature), we can easily adapt the Dual Coordinate Descent (DCD) algorithm 
for our case. 

Following the unconstrained formulation for the SVM objective \citep{hsieh&al08}, 
the MMFS objective in the primal (using a linear kernel) can be written as:
\begin{equation}
\label{eqn:mmfs-dcd}
\min_{w} \quad \frac{1}{2} w^{T}w + \frac{1}{\gamma} \frac{b^2}{2} + C \sum_{i=1}^{N} \xi(w;f_i,r_i)
\end{equation}
where $\xi(w;f_i,r_i)$ denotes the loss function and $\gamma$ is a control parameter. 
Assuming standard $L1$ loss, $\xi(w,x_i,r_i)= \max(r_i-(w_i^{T} f_i+b),0)$. 
Note the slightly changed form of the objective compared to Equation \eqref{eqn:mmfs-primal}
where the bias term $b$ has been replaced by a squared term $\frac{b^2}{2}$. The bias term can 
now be handled by introducing an additional dimension:
\begin{equation}
f'_i \leftarrow [f_i \quad 1/\gamma] \qquad \qquad w'_i \leftarrow [w_i \quad b]
\end{equation}
Equation \eqref{eqn:mmfs-dcd} can then be equivalently written as:
\begin{equation}
\min_{w'} \quad \frac{1}{2} w'^{T}w' + C \sum_{i=1}^{N} \xi(w',f'_i, r_i)
\end{equation}
The dual of this slightly modified problem becomes:
\begin{equation}
\label{eqn:mmfs-dcddual}
\begin{aligned}
f(\alpha) &= \min_{\alpha} ~ \frac{1}{2} \left( \alpha^T Q' \alpha + \gamma * (I^T\alpha)^2 \right) - r'^{T}\alpha \\
\text{subject to} & \quad 0 \leq \alpha_i \leq C, \forall i; 
\end{aligned}
\end{equation}
where $Q'$ is ($N$+$1$)$\times$($N$+$1$) matrix such that $Q'_{ij}={f'}_{i}^{T}f'_{j}$.
Comparing Equation  \eqref{eqn:mmfs-dcddual} 
with Equation \eqref{eqn:mmfs-dual2}, we note that the 
constraint requiring $I^T \alpha=1$ is no longer needed  
because of the slightly changed form of the objective. In the unconstrained form of the dual, we are  
minimizing an additional term $(I^T\alpha)^2$ in the objective which is nothing but the square of the 
$L1$ regularizer over the feature weights. Note that this term in the objective effectively takes care 
of the original constraint $I^T\alpha = 1$. The parameter $\gamma$ controls the 
strength of this regularizer 
and can be tuned to control the sparsity of the solution. The gradient of the objective w.r.t 
to $\alpha_i$ can be computed as follows:
\begin{equation*}
 \begin{aligned}
  {} & G_i = (Q'\alpha)_i +\gamma \sum_{i=1}^N \alpha_i - r'_i
\end{aligned}
\end{equation*}
Using the fact $w'=\sum_{j=1}^N \alpha_j f'_j$ (set of Equations \eqref{mmfs:kkt}), 
the gradient can be further reduced as:
\begin{equation*}
 \begin{aligned}
  {} & G_i = f_{i}^{T}w' +\gamma \sum_{i=1}^N \alpha_i - r'_i
\end{aligned}
\end{equation*}

We adapt the Dual Coordinate Descent algorithm \citep{hsieh&al08} for our 
MMFS problem. This algorithm works by optimizing the 
dual objective by computing the gradient based on the weight vector $w'$ in the 
primal. This process is repeated with respect to each $\alpha_i$ in turn and the weight
vector $w'$ is updated accordingly. This translates into optimizing a one variable quadratic 
function at every step and can be done very efficiently. 
We name this approach MMFS-DCD in the paper, henceforth.
 
\subsection{Complexity}
Following \citep{hsieh&al08}, the MMFS-DCD approach obtains an 
$\epsilon$-accurate solution in $O(\log(1/\epsilon))$ number of
iterations. Time complexity of a single iteration is $O(MN)$. 
Memory complexity of the DCD algorithm is $O(NM)$. For sparse datasets, the 
complexities depend on $\bar{N}$ instead of $N$, where $\bar{N}$ is the average number of 
non-zero feature values in an instance. 
The details about the proof of convergence are available in \citep{hsieh&al08}.
\section{Relationship to Existing Filter Based Methods\label{sec:related}}
Quadratic Programming Feature Selection (QPFS) \citep{lujan&al10} is a filter
based feature selection method which models the feature selection
problem as a quadratic program jointly minimizing redundancy and maximizing
relevance. Redundancy is captured using some kind of similarity score
(such as MI or correlation) amongst the features. Relevance is captured using
the correlation between a feature and the target variable. One norm of
the feature weight vector $\alpha$ is constrained to be $1$. 
Formally, the quadratic program can written as:
\begin{equation}
\label{eqn:qpfs}
\begin{aligned}
f\left(\alpha\right) & = \min_{\alpha} \frac{1}{2}\left(1-\theta\right)\alpha^TQ\alpha - \theta r^T\alpha \\
 \text{Subject to}~~~ & \alpha_i \geq 0, i = 1, ..., N; ~ I^T\alpha = 1.
\end{aligned}
\end{equation} 
$Q$ is an $N\times N$ matrix representing redundancy among the
features, $r$ is an $N$-sized vector representing the feature
relevance and $\alpha$ is an $N$-sized vector capturing feature weights.  
$\theta \in \left[0, 1\right]$ is a scalar which controls the relative 
importance of redundancy (the $Q$ term) and the relevance (the $r$ term).
QPFS objective closely resembles the minimal-redundancy-maximal-relevance 
(mRMR) \citep{peng&al05} criterion. 
When $\theta = 1$, only the relevance is 
considered (maximum Relevance) and when $\theta=0$ only redundancy among 
the features is captured. 
QPFS has also been shown to outperform many existing feature selection methods 
including mRMR and maxRel \citep{lujan&al10}. 

The form of the QPFS formulation above is exactly similar to our dual formulation 
(Equation \ref{eqn:mmfs-dual2}) for an appropriate choice of kernel (similarity) 
function and $C=\infty$ (hard margin). Hence, the QPFS objective falls out as a special 
case of our max-margin framework in the dual problem space when dealing 
with hard margin. It should be noted that Lujan et al. \citep{lujan&al10} do not give any strong
justification for the particular form of the objective used, other than the fact
that it makes intuitive sense and seems to work well in practice. This is unlike our 
case where we present a max-margin based framework for jointly optimizing 
relevance and redundancy. Therefore, our formulation can be seen as providing a 
framework for the use of the QPFS objective and generalizing it further to handle noise
(soft margin). Further, since no direct connection of the QPFS objective has been established 
with the SVM like formulation by Lujan et al. \citep{lujan&al10}, the proposed approach for 
solving the objective is to simply use any of the standard quadratic programming implementations. 
Hence, the time complexity of QPFS approach is $O(N^3 + MN^2)$ and space complexity is 
$O(N^2)$. To deal with cubic complexity, they 
propose combining it with the Nystr\"{o}m 
method which works on subsamples of the data. This can partially alleviate the problem with the 
computational inefficiency of QPFS but comes at the cost of significant loss in accuracy, as 
shown by our experiments.
In our case, because of the close connection with the SVM based max-margin formulation 
and the ability to use the information from the primal as well as the dual, we can utilize any of 
the highly optimized SVM solvers (such as DCD which has time complexity linear in $N$). 

Further it may  be noted that while our MMFS-DCD 
approach can handle sparse representation of 
very high dimensional datasets, other feature 
selection methods like QPFS, FCBF, mRMR etc. cannot 
do so directly.
\section{Experiments}\label{sec:expt}
\label{imple}
\subsection{Datasets}
We demonstrate our experiments on seven publicly available benchmark
datasets with medium to large number of dimensions. Out of these seven datasets Leukemia, RAOA and RAC
are microarray datasets \citep{ganesh12}, MNIST is a vision dataset \citep{mingkui&al10} 
and REAL-SIM, Webspam and Kddb are the text classification 
datasets from NLP domain \citep{chang&al10, yiteng&al12}.
Table \ref{table1} describes the details of 
the datasets. The last column represents 
the sparsity that is average number of non-zero features per instance in the dataset.
\begin{table}[!htb]
\begin{center}
\caption{Dataset description}
\label{table1}
\begin{tabular}{lrrrr}
\hline
{\bf Dataset}  & {\bf \# Training} &{\bf \# Testing} & {\bf \# Features} & {\bf Sparsity} 
 \\
\hline
 Leukemia   &{72} & {-} &  7,129   & 7,129\\ 
 RAOA     	& {31} & {-} & 18,432   & 18,422\\ 
 RAC      	&{33} & {-} &  48,701   & 48,701\\ 
 MNIST		& { 11,982 } & { 1,984 } &{752} & {752}\\ 
 REAL-SIM		& { 57,848 } & { 14,461 } &{20,958} & {51.5} \\
 Webspam  &{80,000}  & {70,000} & {8,355,099} &{3,730 }\\
 Kddb     &{100,000}  & {748,401} & {29,889,813}& {30} \\
\hline
\end{tabular}
\end{center}
\end{table}
\subsection{Algorithms}
We compared the performance of our proposed MMFS algorithm with FCBF\footnote
{http://www.public.asu.edu/\~huanliu/FCBF/FCBFsoftware.html} \citep{yu&liu03}, 
QPFS \citep{lujan&al10} and two other embedded feature selection methods, namely, Feature Generating Machine
(FGM) \citep{mingkui&al10} and Group Discovery Machine (GDM) \citep{yiteng&al12}.
FGM uses cutting plane strategy for feature selection. GDM further tries 
to minimize the redundancy in FGM by incorporating the correlation among the 
features. QPFS, FGM and GDM have been shown to outperform a variety of existing feature selection 
methods including mRMR and MaxRel \citep{peng&al05}, FCBF \citep{yu&liu03}, 
SVM-RFE \citep{guyon&elisseeff03}, etc. 
For QPFS, we used mutual information (MI) as the similarity metric as it has been shown to give 
the best set of results \citep{lujan&al10}. 
In MMFS-DCD, we use correlation of a feature vector with the target class 
vector to compute the feature relevance.
\subsection{Methodology}
We compare all the approaches for feature selection in 
terms of their accuracy and execution time 
on each of the datasets. For all the datasets except 
Webspam and Kddb, we report 
the accuracies obtained at varying number of 
top $K$ features ($K$ = $\{$2, 3, 4,\ldots, 100$\}$) selected for each 
of the methods. For webspam and kddb datasets,we report 
the accuracies obtained at varying number of 
top $K$ features ($K$ =$\{$5, 10, 20, 30,\ldots, 200$\}$) 
selected by FGM, GDM and MMFS-DCD methods.

We also report the best accuracies obtained at any given 
value of $K$ in the above range for all the datasets. 
We have normalized all the datasets except webspam and kddb to zero 
mean and unit variance. The zero mean and unit 
variance normalization for webspam and kddb datasets is very 
memory inefficient (very large memory ($> 100GB$)) as these 
two are very large sparse datasets. We have normalized these 
two datasets with unit variance \citep{yiteng&al12}.
In the microarray datasets, the number of samples are small so 
we report the leave-one-out cross-validation (LOOCV) accuracy.
For MNIST and REAL-SIM datasets, training and testing splits 
are provided in \citep{chang&al10}.
We have followed the training and testing splits of~ \citep{yiteng&al12}
for webspam and kddb datasets. The results reported are 
averaged over $10$ random splits.

For MMFS-DCD, $\gamma$ parameter was tuned separately for each
of the microarray datasets. 
The values of the parameters $C$ and $\theta$ were set to $1$ 
and $0.5$ respectively in all the experiments. 
We used the default settings of the parameters for both FGM and GDM as
reported in \citep{mingkui&al10, yiteng&al12}. After the top $K$ features are selected, 
we used L2-regularized L2-loss SVM \citep{Fan08}
with default settings (that is cost parameter $C$=1) for 
classification for each of the algorithms and for
each of the datasets. MMFS was implemented on top of the 
liblinear tool\footnote{http://www.csie.ntu.edu.tw/~cjlin/liblinear}. 
This implementation uses shrinking strategy \citep{hsieh&al08}.
We used the publicly available implementation of QPFS \citep{lujan&al10}. 
For FGM, we used the
publicly available tool{\footnote{http://www.c2i.ntu.edu.sg/mingkui/FGM.htm}}.
GDM was implemented as an extension of the FGM based on the details given
in Yiteng et. al \citep{yiteng&al12}.
Any additional required wrapper code was written in C/C++. All the experiments 
were run on a Intel Core$^{\text{TM}}$ i7 \@ 3.10GHz machine with 16GB RAM under 
linux operating system.
\subsection{Results}
\label{results}
\subsubsection{Accuracy}
Table \ref{table2} presents the best set of average accuracies (varying the number of 
top-K features selected) for all the methods. 
QPFS method did not produce any results on RAOA and RAC dataset within 24 
hours\footnote{We put a dash $-$ with corresponding entries in the Table \ref{table2}.}. 
So, we used Nystr\"{o}m approximation \citep{lujan&al10} with
sampling rate($\rho$=0.01) for these datasets. 
In the Figure 2(a), 
QPFS-N represents the QPFS with Nystr\"{o}m approximation.
The QPFS and FCBF methods can not handle the sparse data, so we compare 
FGM, GDM and MMFS-DCD for webspam and kddb datasets.
%
MMFS-DCD reaches the best accuracy on 
a small number of top $K$ features for all the microarray datasets. 
Further, MMFS-DCD produces significantly better accuracies compared to FCBF, QPFS, FGM
and GDM on all the microarray datasets (FGM does equally well on RAC). 
On MNIST and webspam datasets, MMFS-DCD is marginally worse than 
the best performing algorithm.
The plots for the average accuracies obtained as we vary the
number of top $K$ features are available in the {\em supplementary} file.
Clearly, for most of 
the datasets, MMFS-DCD is able to achieve the best set of accuracies at early 
stages of feature selection compared to all algorithms. Further, the gene ontology 
and biological significance of top selected genes for leukemia dataset is 
provided in the {\em supplementary} file.
\subsubsection{Time}
\begin{figure*}[!htb]
\centering
 \subfigure[RAC]{\label{fig3c}\includegraphics[width=5.4cm,height=3.77cm]{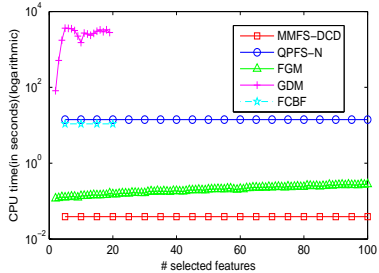}} \hspace{0.01cm}     
  \subfigure[webspam]{\label{fig3e}\includegraphics[scale=0.21]{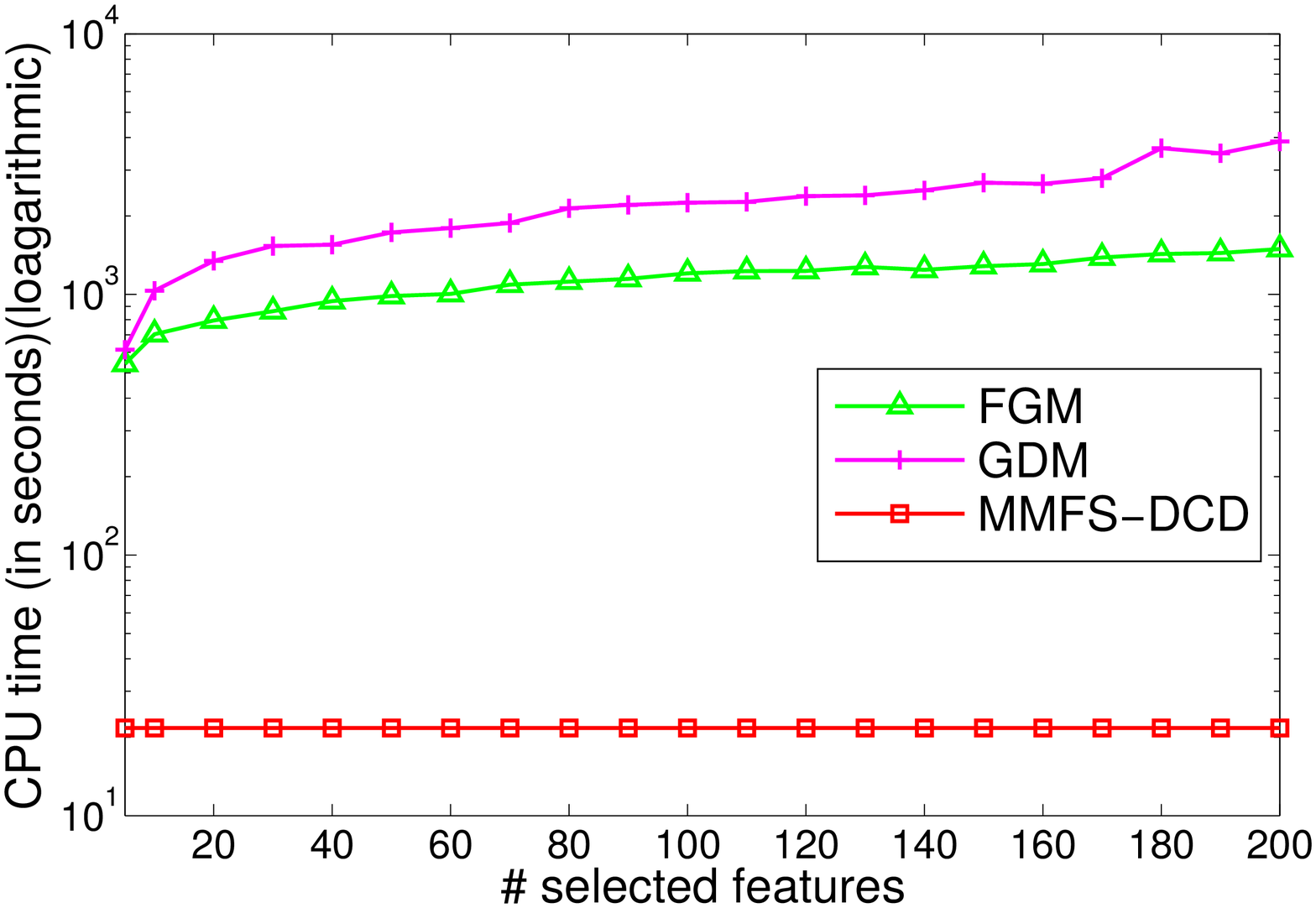}} \hspace{0.01cm}         
  \subfigure[kddb]{\label{fig3f}\includegraphics[scale=0.21]{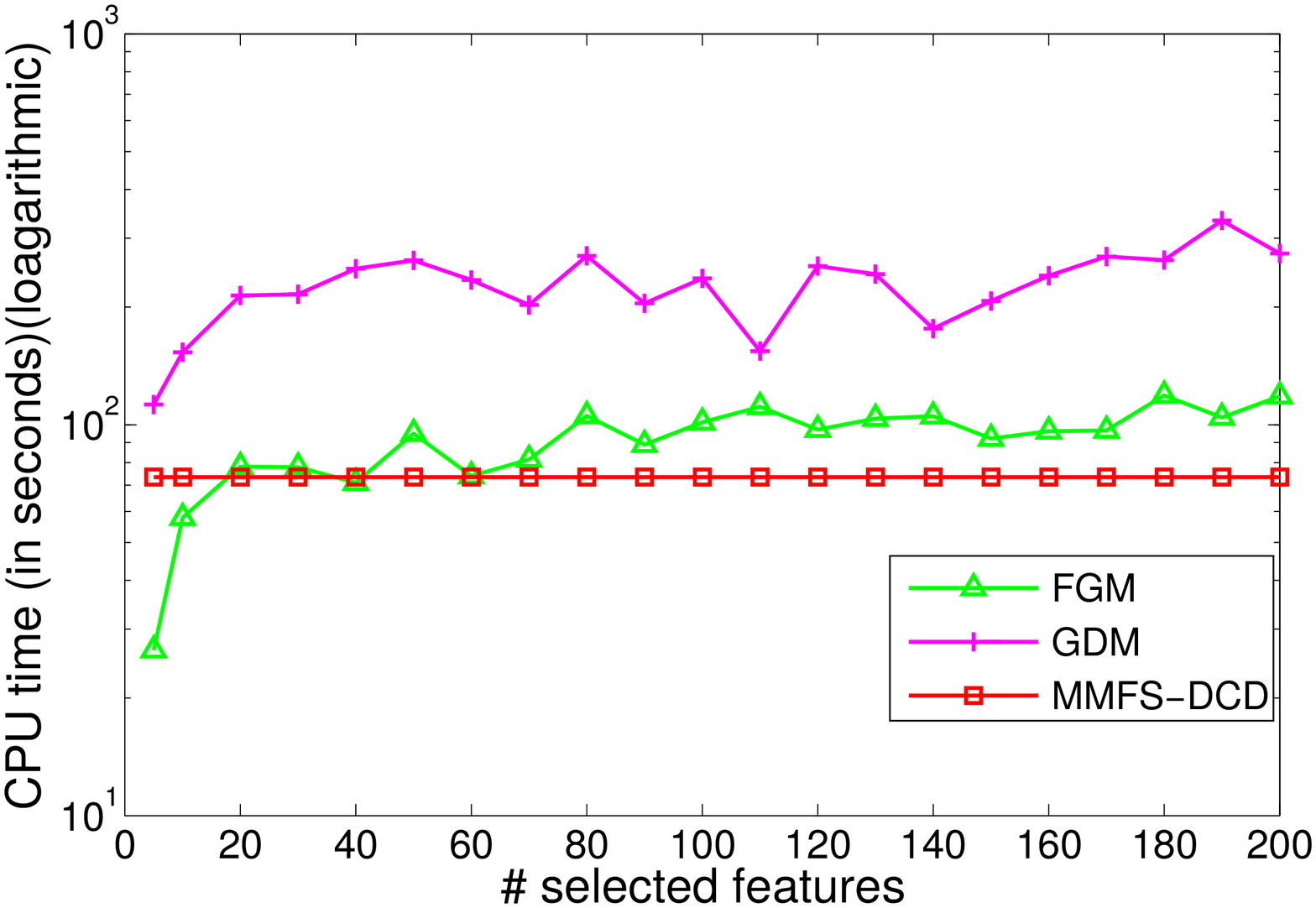}}
\caption{Comparison of execution time (in seconds) of MMFS-DCD with other methods for 
varying number of top $K$ features.}
\label{fig3}
\end{figure*}

Figure \ref{fig3} plots the average execution time for each of the methods.
y-axis is plotted on a log scale. The time requirement for MMFS-DCD, FCBF and QPFS 
is independent of the number of features selected. For FGM and GDM, 
time requirement monotonically increases with $K$. For GDM, there is a sharp 
increase in the time required when $K$ becomes greater than five\footnote{For 
RAC, we run GDM upto $20$ iterations.}. 
It is obvious from Figure \ref{fig3} that MMFS-DCD is upto several orders 
of magnitude faster than all the other algorithms on all the datasets\footnote{
Plots for remaining datasets are available in supplementary file.}. 

\begin{table*}[!htb]
\begin{center}
\caption{Best Accuracy (in \%)}
\label{table2}
\begin{tabular}{lrrrrrrrr{c}r}
\hline
{\bf \quad Dataset}  & \multicolumn{2}{c}{\bf FCBF} & \multicolumn{2}{c}{\bf QPFS} & \multicolumn{2}{c}{\bf FGM} & \multicolumn{2}{c}{\bf GDM}  & \multicolumn{2}{c}{\bf MMFS-DCD}  \\

 &{\bf Accuracy} & {\bf M} &{\bf Accuracy} & {\bf M} & {\bf Accuracy}  &{\bf M}   &{\bf Accuracy} & {\bf M}&  {\bf Accuracy} & {\bf M} \\
\hline
Leukemia   &90.28$\pm$0.1 & 37	 & { 87.50$\pm$ 0.1 } & {45 } & { 87.5$\pm$0.1 } & { 2 } & {84.72$\pm$0.1 } & {2 } &  {\bf 91.67$\pm$0.1} &{ 6 }\\ 
RAOA     	& 74.19$\pm$0.2  &2 & { 67.75$\pm$0.2$^*$ } & { 6 } & { 67.75$\pm$0.2 } & { 2 } &  {54.84$\pm$0.2 } & {2 } &  {\bf 83.87$\pm$0.1} &{ 2 }\\ 
RAC      	& 48.48$\pm$0.2  &12	 & { 96.97$\pm$0.1$^*$ } & {75  } & { 100.0$\pm$0.0 } & { 3 } &  {87.88$\pm$0.1 } & { 3 } &  {\bf 100.0$\pm$0.0 } &{ 2 }\\ 
MNIST	& 91.07$\pm$0.0  &	19& { 96.06$\pm$0.0 } & { 94 } & { 96.21$\pm$0.0 } & {99 } &  { {\bf 96.67$\pm$0.0} } &  { 77  } & { 96.06$\pm$0.0 } &{ 83 }\\ 
REAL-SIM 	& {-}  &{-}	& {- } & { - } & { 90.03$\pm$0.01 } & {90 } &  { 89.48$\pm$0.01 } &  { 100  } & { {\bf 90.19$\pm$0.01} } &{ 100 }\\
Webspam & {-} & {-}   &{-}  &{-} & {95.91$\pm$0.0} &{200} & {\bf 96.80 $\pm$ 0.0}& {200} &{96.79 $\pm$ 0.0} &{200}\\
Kddb    & {-} & {-}   &{-}  &{-} &{87.60 $\pm$ 0.0} & {150} & {87.77 $\pm$ 0.0} &{190} & {\bf 88.39 $\pm$ 0.0} &{200}\\
\hline
\end{tabular}
\end{center}
\end{table*}
\subsubsection{Parameter Sensitivity Analysis}
\begin{figure}[!htb]
\centering
{\includegraphics[scale=0.4]{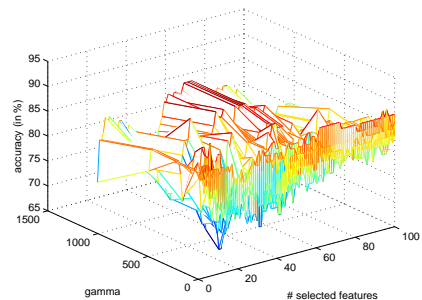}}
\caption{Accuracy variation across $\gamma$ and top $k$ features}
\label{fig4}
\end{figure}

Figure \ref{fig4} presents the variation in accuracy for MMFS-DCD on the
Leukemia dataset, as we vary the regularizer parameter ($\gamma$)
with varying number of top $k$ features. The accuracy is not very 
sensitive to $\gamma$ as demonstrated by a large flat region in 
the graph.
\vspace{-0.1in}
\section{Conclusion and Future Work \label{sec:conc}}
We have presented a novel Max-Margin framework for Feature 
Selection (MMFS) similar to one class SVM formulation. Our framework provides
a principled approach to jointly maximize relevance and minimize redundancy.
It also enables us to use existing SVM based optimization techniques
leading to highly efficient solutions for the task of feature selection.
Our experiments show that MMFS with dual coordinate decent approach is 
many orders of magnitude faster than existing state of the art techniques
while retaining the same level of accuracy.

One of the key future directions includes exploring if there is some notion of 
a generalization bound for the task of feature selection in our framework as in 
the case of SVMs for the task of classification. In other words, what can we say 
about the quality of the features selected as we see more and more data. We would also 
like to explore the performance of our model with non-linear kernels. Lastly, 
exploring the trade-off as we vary the noise penalty would also be a direction to 
pursue in the future.
%
%
%
%
\vspace{-0.1in}
\section*{Acknowledgment}
The authors would like to thank Dr. Parag Singla, Dept. 
of CSE, I.I.T Delhi for his valuable suggestions and support in improving the 
paper.
\bibliographystyle{model2-names}
\bibliography{all}
\end{document}